%% file: main.tex
\documentclass[conference]{IEEEtran}
\IEEEoverridecommandlockouts
\usepackage{cite}
\usepackage{amsmath,amssymb,amsfonts}
\usepackage{algorithm}
\usepackage{algorithmic}
\usepackage{graphicx}
\usepackage{textcomp}
\usepackage{xcolor}
\usepackage{balance}
\usepackage{booktabs}
\usepackage{multirow}
\usepackage{graphicx}
\usepackage{siunitx}
\usepackage{makecell}
\usepackage{bm}

\def\BibTeX{{\rm B\kern-.05em{\sc i\kern-.025em b}\kern-.08em
    T\kern-.1667em\lower.7ex\hbox{E}\kern-.125emX}}
\begin{document}

\title{
Reinforcement Learning as an Improvement Heuristic for Real-World Production Scheduling
\thanks{This work was supported by the Ministry of Economic Affairs, Industry, Climate Action and Energy of the State of North Rhine-Westphalia, Germany, under the project SUPPORT (005-2111-0026)}
}

\author{\IEEEauthorblockN{Arthur Müller} 
\IEEEauthorblockA{\textit{Department of Machine Intelligence} \\
\textit{Fraunhofer IOSB-INA}\\
32657 Lemgo, Germany \\
arthur.mueller@iosb-ina.fraunhofer.de}
\and
\IEEEauthorblockN{Lukas Vollenkemper} 
\IEEEauthorblockA{\textit{Center for Applied Data Science} \\
\textit{Bielefeld University of Applied Sciences and Arts}\\
33330 Gütersloh, Germany \\
 lukas.vollenkemper@hsbi.de}
 }

\maketitle

\begin{abstract}
The integration of Reinforcement Learning (RL) with heuristic methods is an emerging trend for solving optimization problems, which leverages RL's ability to learn from the data generated during the search process.
One promising approach is to train an RL agent as an improvement heuristic, starting with a suboptimal solution that is iteratively improved by applying small changes.
We apply this approach to a real-world multi-objective production scheduling problem.
Our approach utilizes a network architecture that includes Transformer encoding to learn the relationships between jobs. Afterwards, a probability matrix is generated from which pairs of jobs are sampled and then swapped to improve the solution.
We benchmarked our approach against other heuristics using real data from our industry partner, demonstrating its superior performance.
\end{abstract}

\begin{IEEEkeywords}
Reinforcement Learning, Scheduling
\end{IEEEkeywords}

\section{Introduction}
\label{sec:paper6:introduction}
\input{sec/introduction}

\section{Problem Formulation}
\label{sec:paper6:preliminaries}
\input{sec/problem}

\section{Method}
\input{sec/method}

\section{Experiment}
\input{sec/experiments}

\section{Conclusion And Future Work}
\label{sec:paper6:conclusion}
\input{sec/conclusion}

\balance
\bibliography{export}
\bibliographystyle{IEEEtran}

\end{document}

%% file: sec/introduction.tex
Production scheduling (PS) is a problem class from the field of combinatorial optimization commonly encountered in manufacturing.
Adequately solving these problems significantly impacts a company's success, as the quality of the solution greatly contributes to critical metrics such as resource utilization, customer satisfaction and overall costs.

Many production scheduling problems are NP-hard, so that exact solutions within a reasonable time are only possible for small instances, which are usually not practically useful \cite{Zhang2019ProdScheduling,Mamaghan2022}. 
For this reason, approximate algorithms are often used that do not guarantee optimality but are able to find acceptable solutions in reasonable time. One branch of approximate algorithms are problem-dependent heuristics that leverage handcrafted strategies or rules to solve PS problems~\cite{Grumbach2024Diss, Tkindt2006}. Another branch comprises metaheuristics, which are well-studied in the context of PS \cite{Jarboui2013,Hussain2019,Dokeroglu2019}. Metaheuristics are generic optimization methods, often inspired by principles found in nature. In recent years, Reinforcement Learning (RL) has emerged as another promising family of algorithms used in the PS community~\cite{Panzer2022, Mueller2024}.
RL is a type of machine learning where an agent learns to make decisions by performing actions in an environment to maximize cumulative rewards. 

A growing trend is to combine heuristics or metaheuristics with RL to solve optimization problems~\cite{Song2019, Mamaghan2022}.
The combination of these methods is motivated by the fact that traditional heuristics or metaheuristics do not learn from the data generated during their search processes, whereas RL can extract and utilize this data.
For example, RL can learn which operators to use next in a metaheuristic approach~\cite{Mamaghan2023, Zhao2023}. Moreover, RL can directly learn heuristics in a job shop scheduling problem~\cite{Zhang2020}.
Another possibility to combine these methods is to train RL as an improvement heuristic~\cite{Zhang2022,grumbach2024robust}. An improvement heuristic starts with an existing, suboptimal solution and improves it iteratively for a given step limit by applying small changes to the current solution. These changes are usually based on neighbourhood operations, such as swapping elements in a sequence. 
By using RL, it is possible to learn which operations bring about the greatest improvements, which is more effective than conventional metaheuristics, such as simulated annealing, that rely on predefined random strategies without learning from data. According to \cite{Costa2020}, however, only a few studies have dealt with this approach so far, so there is a need for research in this field.  

We contribute to this research field by using RL to learn an improvement heuristic for a real-world, multi-objective production scheduling problem from the automotive sector, demonstrating the applicability of this approach in an industrial context. Our objectives are to minimize the tardiness of jobs in a sequence and the stress on employees.
To create the initial sequence, we use a simple constructive heuristic, namely sorting the jobs according to due dates.
Our network architecture uses Transformer encoding to learn the relationships between jobs and generates a probability matrix from which a job pair is sampled. The jobs in this pair are then swapped. This process is repeated multiple times to improve the initial sequence according to the optimization objectives. 
To evaluate our method, we compared its performance with other heuristics, using real-world data provided by our industry partner. The results show that our approach outperforms the other methods.

The strength of our approach furthermore lies in its flexibility: a single network can handle sequences with varying numbers of jobs and can integrate different operators besides swapping. Additionally, it can be combined with various constructive heuristics that generate the initial solution. This flexibility makes it suitable for a wide range of real-world applications.

%% file: sec/problem.tex

\begin{figure}
    \centering
    \includegraphics[width=0.48\textwidth]{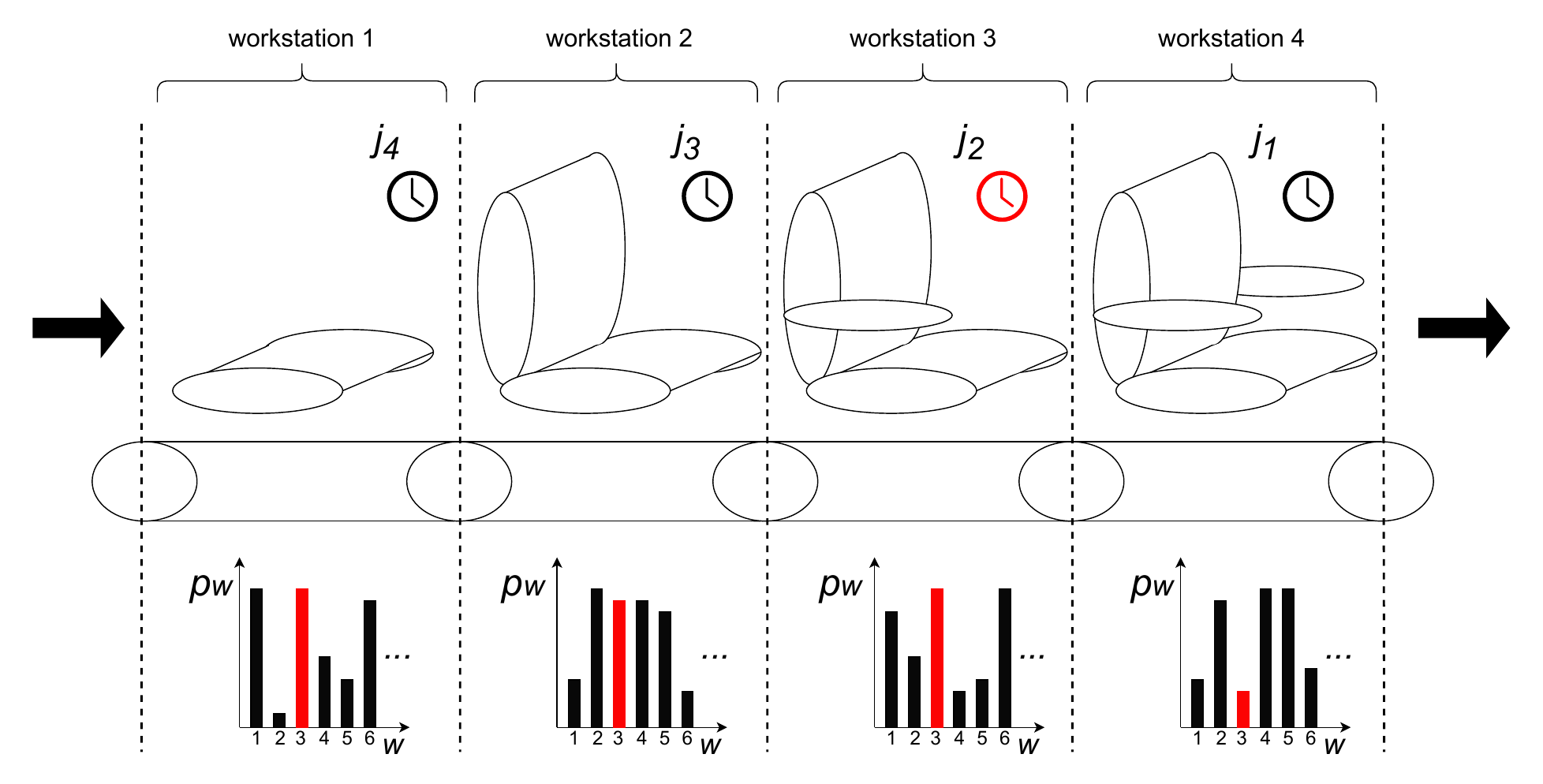}
    \caption{Simplified illustration of the production line. In this example, seat $j_2$ is not completed by the due date. In addition, there is stress for the employee at workstation 3, as the consecutive seats $j_2$ to $j_4$ at this station have long processing times. The sequence of seats could be improved by swapping seats $j_2$ and $j_1$, as $j_1$ has a low processing time at workstation 3 and the tardiness of $j_2$ would be reduced.}
    \label{fig:paper6:production}
\end{figure}

The present problem is derived from a real-world production scheduling problem of an automotive supplier in Germany, 
presented in \cite{Vollenkemper2023}. 
The company produces customized car seats and delivers them just-in-sequence to its customers. 
Seats are manufactured in a production line with several workstations, each performing different operations.
All seats move at the same speed and maintain equal distances from each other.
As a result, the same amount of time is available at each workstation to complete an operation. 
However, the actual time to complete an operation at a workstation, denoted as processing time, is less than or equal to this time window and varies depending on the seat configuration. The processing times of all operations are known based on systematic measurement during actual production (Methods Time Management).
Furthermore, each seat has a due date by which it must be produced, and the seats cannot overtake each other on the production line. An illustration of the production line is shown in Figure~\ref{fig:paper6:production}.

For a given set of seats to be produced, we aim to find a permutation that optimizes two objectives:
\begin{enumerate}
    \item  The tardiness of all jobs - the time required for completion beyond the due date - should be kept to a minimum. This objective is intended to ensure the economic efficiency of the production.
    \item The second objective addresses the well-being of employees: stress for employees should be kept to a minimum. Stress results from consecutive seats with high processing times at a workstation, as this results in less buffer and recovery time for the employees. Conversely, consecutive seats with short processing times at a workstation should also be avoided because this could lead to boredom, which is another kind of stress~\cite{van2014boredom,wan2014understanding}. Therefore, the absolute difference between all processing times of consecutive seats should be maximized. 
\end{enumerate}
This scheduling problem can be formalized as a so-called Permutation Flow Shop Problem. In the following, we provide a mathematical problem definition.


We have $N$ seats, which we refer to as jobs in the following, and $W$ workstations. A job $j$ is defined as a tuple $(p^j_1,\dots,p^j_W, d_j)$, where $p^j_w \in \mathbb{R}^+_0$ corresponds to the processing time of that job on workstation $w$ and $d_j \in \mathbb{R}^+$ to the due date. We refer to the set of all jobs as $J=\{j_1,\dots,j_N\}$. A solution to the optimization problem is a permutation $\sigma$ of $J$, where $\sigma(i)$ denotes the job index of the $i$-th job in $\sigma$.

The time available for completing an operation on a workstation, $T_\text{W} \in \mathbb{R}^+$, is the same for all jobs at all workstations.
Therefore, each processing time cannot be greater than $T_\text{W}$:  
\begin{equation}
    0 \leq p^{j}_{w} \leq T_\text{W}. 
\end{equation}
A job is completed when all predecessor jobs have been released and the job has passed all workstations. The completion time $C_i$ of a job at the $i$-th position can therefore be calculated by 
\begin{equation}
    C_i = T_\text{W}(W+i-1).    
\end{equation}
We define the tardiness $T_{\text{T}}$ of the $i$-th job in the permutation $\sigma$ to be 
\begin{equation}
    T_{\text{T}}(\sigma,i) = C_i-d_{\sigma(i)}.
\end{equation}

Our first objective $f_1$ is minimizing the tardiness of all jobs. Instead of using tardiness directly, we weight it with the exponential function $g_{\text{T}}$:
\begin{equation}
    g_{\text{T}}(\sigma,i) = e^{T_{\text{T}}(\sigma,i)}.
\end{equation}

By doing so, stronger emphasis is put on high tardiness, while all punctual jobs ($T_{\text{T}}(\sigma,i) \leq 0$) are roughly treated equally.
The first objective can then be formalized as:
\begin{equation}
    \label{eq:paper6:obj1}
    f_1(\sigma) = \sum_{i=1}^{N} g_{\text{T}}(\sigma,i).
\end{equation}

The second objective $f_2$ is to ensure that small and large processing times for a workstation alternate between consecutive jobs so that the employees on this workstation experience as little stress as possible. This can be formalized as 

\begin{equation}
    \label{eq:paper6:obj2}
    f_2(\sigma) = \sum_{w=1}^{W} \sum_{i=1}^{N-1} |p^{\sigma(i)}_{w}-p^{\sigma(i+1)}_{w}|,
\end{equation}
where $f_2(\sigma)$ is to be maximized.

An illustration of how the objectives relate to a given permutation is shown in Figure~\ref{fig:paper6:production}.

%% file: sec/method.tex
\subsection{Improvement Heuristic}
We utilize the RL agent as an improvement heuristic. In this approach, an initial solution $\sigma_0$ at step $t=0$ is generated by an algorithm, such as a simple heuristic. Starting from this initial solution, a local operator modifies the current solution $\sigma_t$ at each iteration, producing a new solution $\sigma_{t+1}$.

\begin{figure}
    \centering
    \includegraphics[width=0.48\textwidth]{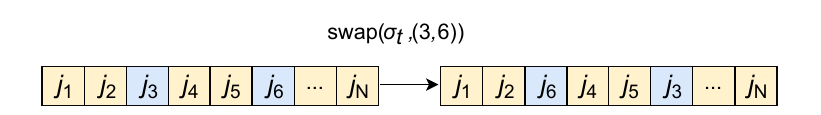}
    \caption{Illustration of the $\text{swap}$-operator. $\text{swap}(\sigma_t, (3, 6))$ exchanges the positions of the third and 6th job in the permutation $\sigma_t$.}
    \label{fig:paper6:swap}
\end{figure}

The local operator modifies $\sigma_t$ to better achieve the objectives. In our approach, we consider only pairwise operators, which are commonly used in the context of production scheduling. Classic examples include $\text{swap}$, $\text{shift}$, and $\text{insert}$. With $\text{shift}$, a job in the permutation is moved one position forward or backward, while with $\text{insert}$, a job is relocated from its current position to another position in the permutation. In our experiments, we use the swap operator. The function $\text{swap}(\sigma_t, (i, k))$ exchanges the positions of the $i$-th and $k$-th job in the permutation $\sigma_t$, see Figure~\ref{fig:paper6:swap} for an illustration.

We generate the initial permutation by sorting all jobs according to their due dates, thereby optimizing $f_1$ (tardiness). However, in most cases optimizing $f_1$ is in conflict with optimizing $f_2$ (stress). To balance both objectives, we evaluate the objectives of new permutations relative to the initial permutation and weight them:
\begin{equation}
    \Delta f_1(\sigma_t) = f_1(\sigma_0) - f_1(\sigma_t),
\end{equation}
and
\begin{equation}
    \Delta f_2(\sigma_t) = f_2(\sigma_t) - f_2(\sigma_0).
\end{equation}
Note that $\Delta f_1(\sigma_t) \leq 0$, since $f_1(\sigma_0)$ is already the minimum.
The combined, weighted objective $f_\text{c}$ is then given by
\begin{equation}
    \label{eq:paper6:objcombined}
    f_\text{c}(\sigma_t) = \alpha_1\Delta f_1(\sigma_t) + \alpha_2\Delta f_2(\sigma_t),
\end{equation}
where $\alpha_1$ and $\alpha_2$ represent the weights and are also used to normalize the objectives to the same range. The objective should be maximized so that stress is minimized as much as possible while the weighted tardiness does not become too large.

The RL agent is allowed to run for a fixed number of steps $T$. Afterwards, the best solution $\sigma^*$ found during the run is returned.

\subsection{RL Formulation}
In the following we are formulating the Markov decision process (MDP) for the RL agent to act as an improvement heuristic that swaps jobs at each iteration.

\subsubsection{state}
We define the state $s_t$ at step $t$ as a sequence of feature vectors $x(j_i)$, for each job $j_i$ in the permutation $\sigma_t$. The state $s_t$ retains the order of $\sigma_t$.


\subsubsection{action}
The RL agent should apply the swap operator to the current permutation at each step. Therefore, we define the action $a_t$ as a pair of positions $(i, k)$ in the permutation that indicate which jobs to swap. The new permutation is then obtained by $\sigma_{t+1} = \text{swap}(\sigma_t, (i, k))$.


\subsubsection{reward}
We aim to maximize the combined objective in Eq.~\ref{eq:paper6:objcombined} relative to the number of steps. Therefore, we define the reward function as
\begin{equation}
    r_t = f_\text{c}(\sigma_{t+1})/T.
\end{equation}
The cumulative reward over an episode (return) is given by:
\begin{equation}
    G_T = \sum_{t=0}^{T-1}\gamma^t r_t = \frac{1}{T}\sum_{t=0}^{T-1}\gamma^t f_\text{c}(\sigma_{t+1}).
\end{equation}
By using the discount factor $\gamma \in [0, 1]$, it is possible to prioritize immediate rewards over future rewards.

However, this definition has a potential issue. The agent might avoid temporarily worse solutions that could lead to better overall outcomes, as these temporary worse solutions would lower the overall return. That is the reason why we initially defined a reward function similar to the one proposed in \cite{Wu2022}. In that reward function, we considered the difference between the best value achieved so far $f_\text{c}(\sigma^*_t)$ and the current value $f_\text{c}(\sigma_{t+1})$ if the current value was better. Otherwise, the reward function returned 0. This ensured that temporarily worse solutions did not affect the return. $G_T$ would reflect only the improvement from $f_\text{c}(\sigma_0)$ to $f_\text{c}(\sigma^*_T)$.
However, we found empirically that this reward function performed worse in our case. A reason for this could be its sparsity, which makes training harder.

\subsubsection{policy}
The decisions $a_t$, determining which jobs to swap, are made by a stochastic policy $\pi$. In total, the policy is allowed to make $T$ swaps. The transition process is described by the probability chain rule:

\begin{equation}
    P(s_T|s_0) = \prod_{t=0}^{T-1} \pi({a_t|s_t}).
\end{equation}

Note that we use a stochastic policy both during training (for exploration) and inference. Empirically, this approach has proven to be more effective than using a deterministic policy, which is consistent with the findings in \cite{Wu2022}. 

\subsection{Network Architecture}
\begin{figure*}
    \centering
    \includegraphics[width=0.99\textwidth]{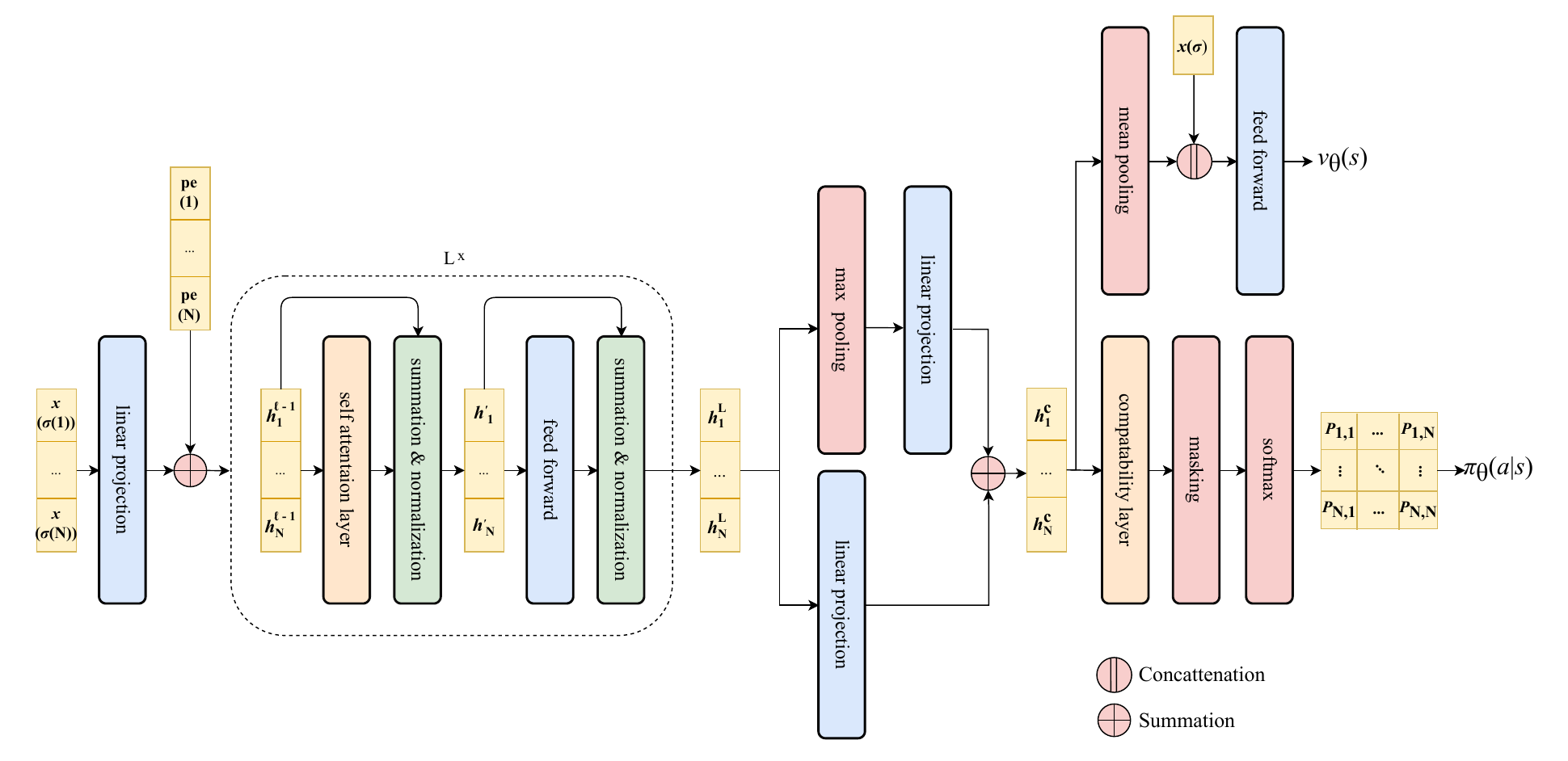}
    \caption{Architecture of the network used, consisting of a policy and critic network with shared parameters. The network employs Transformer encoding to process the features of the permutation $\sigma$. The policy network calculates a probability matrix, from which the next job pair to be swapped is selected ($\pi_\theta(a|s)$). The critic network estimates the remaining cumulative reward $v_\theta(s)$, which is used as information during training by the PPO algorithm. During inference, only the policy network is utilized.}
    \label{fig:paper6:nn_architecture}
\end{figure*}

We use the Proximal Policy Optimization (PPO) algorithm \cite{Schulman2017} to learn the policy.
For this reason, we parameterize the policy $\pi_\theta$ as an artificial neural network with the trainable parameters $\theta$. The network is illustrated in Figure~\ref{fig:paper6:nn_architecture}.
Furthermore, the network comprises a critic network, which is used to estimate the remaining cumulative reward from any given state $s$, denoted as $v_{\theta}$. Note that the policy network and critic network share parameters.

The policy network consists of two parts. The first part uses Transformer encoding \cite{Vaswani2017} to process the features of the jobs and represent their relationships and interactions within the permutation. We chose this approach because Transformer encoding, due to the self-attention mechanism, have proven to be very useful in recent years for learning dependencies and interactions between elements in sequences \cite{Niu2021,Lin2022}.
Furthermore, the policy network is flexible in terms of the number of jobs per permutation. This means that the same policy network can process permutations of different lengths, which is a significant advantage over the use of simpler nets, such as multilayer perceptrons.

The second part selects the job pair to be swapped. Inspired by \cite{Wu2022}, we chose a compatibility layer for job pair selection that is adopted from \cite{Vaswani2017}. This calculates a probability matrix, representing the probability of each job pair to be selected.
In the following, we will describe the network in more detail:

\subsubsection{Job Embedding}
For better readability, the step index $t$ is omitted in the following. The input to the policy network are the features of each job $\{x(\sigma(i))\}_{i=1}^N$ of a permutation. 
First, we increase the dimensions of the job features to a higher-dimensional embedding space $h^0_i$ with dimensions $d_h$ using a linear transformation. This enhances the model's capacity to capture complex patterns and relationships within the data. 
We set $d_h=128$, as this has been empirically shown to be well suited in previous work \cite{Wu2022,Vaswani2017}.  
Furthermore, we add sinusoidal positional encodings $\text{pe}(i)$ to $h^0_i$, resulting in $h^0_i =h^0_i + \text{pe}(i)$, to capture the positions of the jobs relative to each other.

Next, the embeddings are further processed by the Transformer encoding layers: 
\begin{equation}
    h_i' = \textbf{LN}^{l,1}(h_i^{l-1} + \textbf{MHA}^l_i(h^{l-1}_i,\ldots, h_{N}^{l-1})),    
\end{equation}
where $\textbf{MHA}$ represents a multi-head attention layer, $l$ represents the layer index, and $\textbf{LN}^{l,1}$ the first of two layer normalization operators in layer $l$.
We chose the number of heads in $\textbf{MHA}$ to be $2$, as this has empirically shown to perform best. The dimension of the embeddings $h_i'$ is $d_h$. 



Afterwards, the embeddings are further transformed through a feed forward layer ($\textbf{FF}_\textbf{p}$):
\begin{equation}
    h_i^l = \textbf{LN}^{l,2}(h_i' + \textbf{FF}_\textbf{p}^l(h_i^l)),
\end{equation}
where $\textbf{LN}^{l,2}$ denotes the second layer normalization operation in layer $l$. $\textbf{FF}_\textbf{p}$ has a hidden layer with a dimension of $512$ followed by a ReLU activation function. We chose the numbers of encoding layers to be two, as this shown be sufficient for processing the job features. Each encoding layer has its own set of trainable parameters.   

The resulting embeddings $\{h_i^2\}_{i=1}^N$ are aggregated by max-pooling to extract global information about the permutation:
\begin{equation}
    h_\text{max} = \max \{h_i^2\}_{i=1}^N.
\end{equation}
This global information are then integrated back to each embedding by a linear transformation, as this was done e.g. in \cite{Song2023} and \cite{Wu2022}, resulting in $h_i^c$:
\begin{equation}
    h_i^c = W^i h_i^2 + W^\text{max} h_\text{max},
\end{equation}
where $W^i, W^\text{max} \in \mathbb{R}^{d_h \times d_h}$.

\subsubsection{Job Pair Selection}
Based on the job embeddings $H^c = [h_1^c, h_2^c, \ldots, h_N^c]$, our goal is to determine which job pair to swap in order to improve the permutation. For this purpose, we use a compatability layer inspired by the attention mechanism \cite{Vaswani2017}. The embeddings are multiplied with the so called query and key matrices $Q_c = W_c^q H^c$ and $K_c = W^k_c H^c$ with $W_c^q, W_c^k \in \mathbb{R}^{d_h \times d_h}$. Afterwards, the compatability matrix $Y$ is calculated by 
\begin{equation}
    Y = K_c^T Q_c, \quad Y \in \mathbb{R}^{N \times N}.
\end{equation}

Each element in the matrix $Y_{ik}$ indicates how suitable a pair of positions $(i, k)$ is for the next swap. Furthermore, we mask out the main diagonal of $Y$ in order to avoid swapping identical jobs:
\begin{equation}
    \tilde{Y}_{ik} = \begin{cases}
    \text{ReLU}(Y_{ik}), & \text{if } i \neq k \\
    -\infty, & \text{if } i = k
    \end{cases}
\end{equation}
Finally, we apply softmax to obtain the probability matrix $P = \text{softmax}(\tilde{Y})$ with which the job pairs are sampled by the policy.



\subsubsection{Critic Network}
The critic network estimates the cumulative reward $v_\theta(s)$ at each step, as PPO requires this information during training. For this purpose, we aggregate the job embeddings $H^c$ using mean pooling into a single vector $h_\text{mean} = \text{mean} \{h_i^c\}_{i=1}^N$. This vector is then concatenated with the general features of the permutation $x(\sigma)$, which has the dimension $d_\text{gen}$.
We found that including $x(\sigma)$ significantly improves the accuracy of the estimation.
The concatenated vector is then fed into a feed-forward network $\textbf{FF}_\textbf{v}$ with two layers, each having $d_h$ neurons, to estimate $v_\theta(s)$. The input dimension of $\textbf{FF}_\textbf{v}$ is $d_h + d_\text{gen}$ and the output dimension $1$:  
\begin{equation}
    v_\theta(s) = \textbf{FF}_\textbf{v}(\text{Concat}(h_\text{mean},x(\sigma)))    
\end{equation}

\subsection{Job Features}
\label{sec:paper6:features}
To enable the policy of swapping jobs to optimize the objective, we need to include informative features. For the $i$-th job in $\sigma$, we define the feature vector to be: 
    $x(\sigma, i) = (p^{\sigma(i)}_{1}, \dots, p^{\sigma(i)}_{W}, p^{\sigma(i)}_{1}-p^{\sigma(i+1)}_{1}, \dots, p^{\sigma(i)}_{W}-p^{\sigma(i+1)}_{W}, d_{\sigma(i)}, g_{\text{T}}(\sigma,i))$.
Note that we do not use the absolute difference between processing times as in Eq.~\ref{eq:paper6:obj2}, but rather the direct difference to provide the policy with information about the direction of deviation. 
The differences for the last job $N$ in the permutation are set to 0.
We also include both the due dates and the weighted tardiness.
Thus, the feature vector has a dimension of $2 \times W + 2$.


We define the general feature as $x(\sigma)=t/T$. This information is necessary for the critic network to estimate how much reward can still be collected within the remaining steps. Contrary to our intuition, including $x(\sigma)$ also as information for the policy network did not improve performance, so we excluded it from the final version of the network architecture.






%% file: sec/experiments.tex
To evaluate our approach, we conducted experiments with real-world data from our industry partner Isringhausen and compared the performance against other heuristics.

\subsection{Experimental Setup}
The data set comprises $396$ sets of jobs, each consisting of $N=20$ jobs that are processed by $W=12$ workstations. Each job has $2\times W+2=26$ features. 
The time for completing an operation at a workstation is $T_W=\SI{208}{\second}$.
We used $200$ job sets for training and held back the remaining $196$ for testing. We set the number of job pair swaps (i.e. the number of steps per episode) to $T=10$ and the number of training steps to $2 \times 10^6$. At the start of each episode, a job set was randomly sampled from the training data. 

\subsection{Training and Inference}

For training and evaluation, we have implemented an environment in Python that enables an efficient calculation of job features and rewards and the execution of the swap operator without having to rely on specialized simulation tools such as SimPy. The network was implemented in PyTorch~\cite{Pytorch2} and has a total of $482817$ trainable parameters. For training the network, we used the PPO implementation from ray rllib \cite{Liang2018} as this framework allows for efficient parallelization. The following hyperparameters for PPO were empirically set (following the notation of ray rllib): $\texttt{clip\_param}=0.2$, $\texttt{gamma}=0.99$, $\texttt{lambda}=0.99$, $\texttt{train\_batch\_size}=1024$, $\texttt{sgd\_minibatch\_size}=32$, $\texttt{num\_sgd\_iter}=20$. The learning rate was linearly reduced from $5 \times 10^{-4}$ at the beginning of training to $2 \times 10^{-5}$ at the end of training to improve convergence.

Although training this agent is time-consuming, the inference is computationally relatively cheap and fast.
We take advantage of this and run the agent, as in \cite{Wu2022}, with two diversifying strategies, namely \textit{multirun} and \textit{multipolicy}.
In the multirun strategy, we use the final policy, which was saved after the completion of training, and run it $30$ times with a stochastic policy. The best permutation is kept. We refer to this policy as RL-MR. 
Stochastic policy in our case means that the next job pair to be swapped is sampled from the probability matrix $P$ rather than selecting the pair with the highest value, as would be done in a deterministic policy. This allows for a better exploration of the solution space.
In the multipolicy strategy, we use the final policy along with five earlier policies. It is possible that these policies reach different regions in the solution space that yield better performance for some permutations. We combine multipolicy with multirun, so that each of these policies is also run $30$ times with a stochastic policy, denoted as RL-MPMR. This results in a total of 300 steps for RL-MR and 1800 steps for RL-MPMR.


\subsection{Comparison with other heuristics}
To better assess the performance of our approach, we benchmarked it against two other heuristics. First, we used Simulated Annealing (SA) \cite{Kirkpatrick1983}, a well-established metaheuristic inspired by the annealing process in metallurgy. 
We utilized the implementation provided by the Python package simanneal\footnote{https://github.com/perrygeo/simanneal}. Starting from a due date sorted permutation, random job pairs are swapped to improve it. To avoid getting stuck in local optima, worse permutations are temporarily accepted during the optimization process. 
The probability of accepting worse permutations is determined by the so-called temperature parameter, where higher values increase the probability of accepting worse solutions. 
Starting at $\texttt{Tmax}$, the temperature decreases exponentially during the optimization process and eventually is reaching a minimum value $\texttt{Tmin}$.

We evaluated SA with three different parameter settings. In the SA-300 and SA-1800 settings, the number of steps was limited to 300 and 1800, respectively, matching the number of steps used by RL-MR and RL-MPMR during inference. Additionally, we tested a variant where SA was allowed to perform 530,000 steps (SA-530k), which, on our hardware, corresponds to a runtime of approximately 1 minute. Longer runtimes were considered impractical for real-world deployment in this case. The $\texttt{Tmax}$ and $\texttt{Tmin}$ values were determined using the $\texttt{auto}$ method of the simanneal package, resulting in values of $72.0$ and $2.2e-61$, respectively.       

Secondly, we developed a simple heuristic (SH), which also starts from a permutation sorted by due date $\sigma_0$. Initially, the first element of $\sigma_0$ is selected as the first element in the final permutation. Then, the following process is repeated until the final permutation is complete: From the last scheduled job $i$, the job $i'$ that maximizes the distance according to Eq.~\ref{eq:paper6:obj2} is chosen from the next $n_{\text{SH}}$ jobs:
\begin{equation}
i' = \arg\max_{k} \sum_{w=1}^W |p^{\sigma(i)}_{w}-p^{\sigma(k)}_{w}|, 
\quad k \in [i+1,i+n_{\text{SH}}]
\end{equation}
Furthermore, each time a job from the current subset is not selected as next job, a skip counter is incremented for this job. If this counter exceeds a predefined threshold $\texttt{max\_skip}$, the corresponding job is scheduled next, even if it does not maximize the distance.
This is to prevent the tardiness of individual jobs from becoming too high. We parameterized this heuristic with different settings to generate a variety of benchmark solutions. For example, we refer to SH parameterized with $n_{\text{SH}}=4$ and $\texttt{max\_skip}=4$ as SH-n4ms4.


\subsection{Results and Discussion}
Results are shown in Table~\ref{tab:paper6:evaluation_metrics}.
The combination of multipolicy and multirun achieve the best results by far. For instance, $f_c$ on the training data is $31\%$ higher with RL-MPMR than with RL-MR, and on the test data, it is even $38.4\%$ higher.
This can also be seen in the objectives $f_1$ and $f_2$. For example, RL-MPMR was able to improve $f_2$ by $2.4\%$ compared to RL-MR in the test data, whereas only an increase of $0.3\%$ had to be tolerated for $f_1$.
Thus, saving policies at different stages of training proves beneficial, as they can sometimes find better permutations than the final one.


\begin{table*}[ht]
\centering
\caption{Comparison of evaluation metrics for different heuristics on the train (200 job sets) and test (196 job sets) set. The optimization objective $f_c$ is emphasized. The optimization objectives $f_c$, $f_1$, and $f_2$ are averaged over all train and test job sets, respectively. Additionally, the number of job sets in which no improvement could be achieved is listed.}
\label{tab:paper6:evaluation_metrics}
\begin{tabular}{llcccccccccc}
 & \makecell{evaluation \\ metrics}  & RL-MR & RL-MPMR & SA-300 & SA-1800 & SA-530k & \makecell{SH-n4ms4} & \makecell{SH-n4ms6} & \makecell{SH-n6ms8} & \makecell{SH-n6ms10} & \makecell{SH-n8ms10} \\
\hline
\multirow{4}{*}{\rotatebox[origin=c]{90}{Train}} & \(\bm{f_c}\) & \textbf{6.96} & \textbf{9.12} & \textbf{1.82} & \textbf{2.75} & \textbf{6.23} & \textbf{1.75} & \textbf{2.12} & \textbf{0.8} & \textbf{0.99} & \textbf{0.23} \\
 & \(f_1\) & 11.90 & 11.91 & 12.12 & 12.33 & 12.44 & 11.75 & 11.83 & 12 & 12.13 & 12.25 \\
 & \(f_2\) & 10134 & 10322 & 9722 & 9899 & 10292 & 9433 & 9522 & 9310 & 9352 & 9097 \\
 & \# no impr. & 1 & 0 & 89 & 62 & 6 & 82 & 63 & 120 & 126 & 166 \\
\hline
\multirow{4}{*}{\rotatebox[origin=c]{90}{Test}} & \(\bm{f_c}\) & \textbf{5.76} & \textbf{7.97} & \textbf{2.0} & \textbf{3.34} & \textbf{6.68} & \textbf{1.38} & \textbf{1.87} & \textbf{1.28} & \textbf{1.45} & \textbf{1.1} \\
 & \(f_1\) & 11.54 & 11.58 & 11.86 & 12.06 & 12.15 & 11.4 & 11.48 & 11.63 & 11.76 & 11.89 \\
 & \(f_2\) & 9821 & 10060 & 9573 & 9788 & 10156 & 9164 & 9307 & 9200 & 9232 & 9094 \\
 & \# no impr. & 2 & 0 & 76 & 39 & 2 & 98 & 87 & 118 & 109 & 136 \\
\hline
\end{tabular}
\end{table*}

Furthermore, our RL method significantly outperforms SA when both are allowed the same number of steps. For instance, RL-MR achieves approximately 3.8 times the $f_c$ value of SA-300 on the training data, and still about 2.8 times on the test data. Similar results are observed when comparing RL-MPMR and SA-1800, where $f_c$ is about 3.3 times higher for RL-MPMR on the training data and 2.4 times higher on the test data. Additionally, SA-300 and SA-1800 exhibit a significantly high number of permutations that do not improve over the initial permutation, with a total of $165$ and $101$, respectively. In contrast, for RL-MR, this is only $3$, and for RL-MPMR, it is even $0$.

However, when massively increasing the number of steps, we observe that the performance of SA approaches that of RL-MR. A comparison between RL-MR and SA-530k shows better results for RL-MR on the training data and for SA-530k on the test data. Notably, SA-530k prioritizes $f_2$ and even slightly outperforms RL-MPMR on the test data, achieving $10156$ compared to $10060$. However, this comes at the expense of tardiness ($f_1$), where both RL variants outperform SA-530k on both the training and test data.

As for the SH variants, while they are computationally the most efficient, they perform the worst, with SH-n8ms10 showing the lowest performance.



Another insight from the results is the decrease in performance on the test data for our approach, indicating that the network does not generalize sufficiently.
This assumption is also supported by the finding mentioned above that in 3 cases RL-MR does not bring any improvement compared to due date sorting.
The probably cause is the lack of sufficient training data, which is typically required for large networks like ours.
To address this, in future work we want to collect additional data as well as generate synthetic data to improve the generalization ability of the policy.

\subsection{Visualization}
\begin{figure*}
    \centering
    \includegraphics[width=0.68\textwidth]{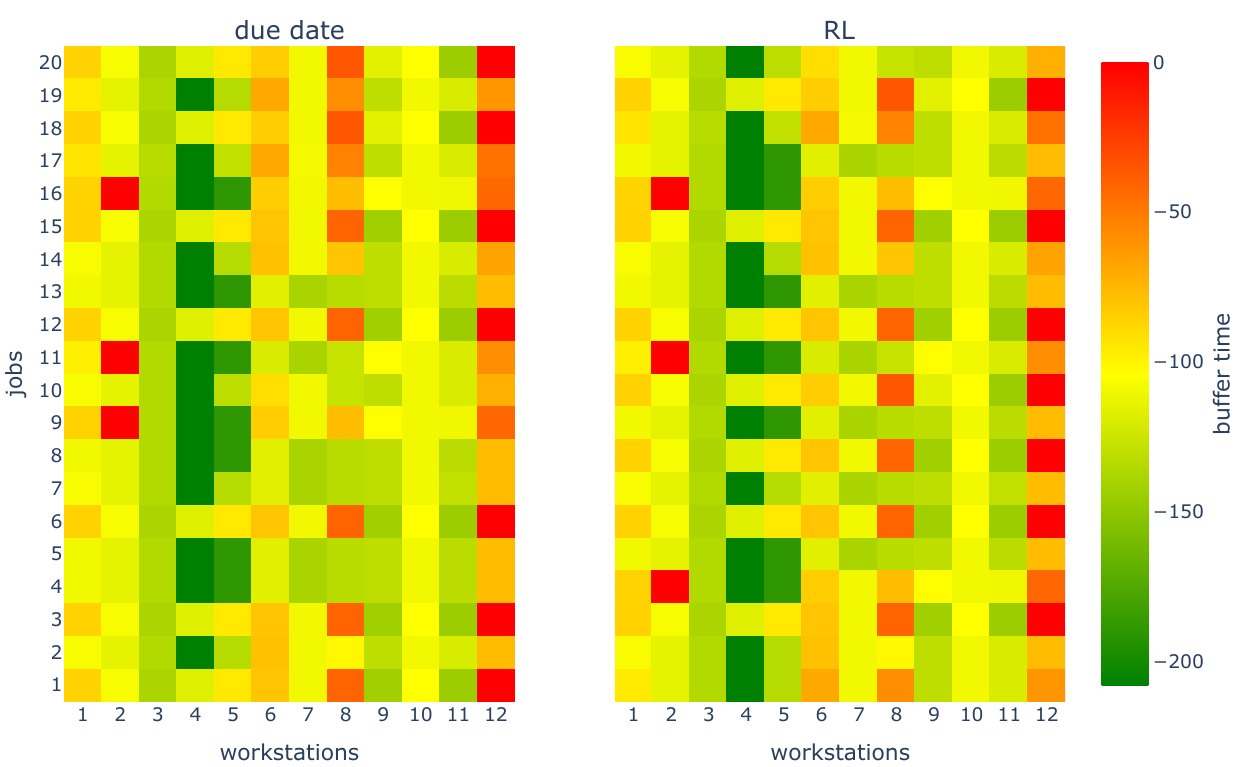}
    \caption{The heatmap shows the buffer time $p^j_{w}-T_\text{W}$ of all operations in an example permutation of the train data. On the left, the permutation is sorted by due date. On the right, the permutation is shown after 10 swaps by the RL agent. Green indicates more buffer time, red less. It can be seen, that in the right permutation long and short operations alternate more often.}
    \label{fig:paper6:plans}
\end{figure*}

To better understand the behavior of the learned policy, we visualize a specific job set from the training data (see Figure~\ref{fig:paper6:plans}).
The heatmaps shown in this figure show the differences between the processing times of all operations and the time available $T_\text{W}$ to complete an operation $p^j_{w}-T_\text{W}$, referred to as buffer time in the following. Green indicates more buffer time, red less. On the left, the permutation is sorted by due date. On the right, the permutation is shown after 10 swaps by RL-MR.
It is evident that the jobs have been reorganized in such a manner that, for example, operations on workstation 4 with a lot of buffer time now alternate more frequently with jobs that have less buffer time. This can also be observed on workstation 8, where jobs with less buffer time now alternate more frequently with jobs that have more buffer time.
This behavior is desired, as it reduces stress for employees.

%% file: sec/conclusion.tex
In this paper, we trained an RL agent as an improvement heuristic to solve a real-world, multi-objective production scheduling problem. To this end, we developed a network architecture that employs Transformer encoding to capture the relationships between jobs. The network can process varying lengths of permutations. Based on the job embeddings, a probability matrix is generated from which a pair of jobs is selected and swapped to incrementally improve the permutation. We tested our method against other heuristics using real-world data from our industry partner, and it proved to be superior.

In future, we plan to extend this approach to several production lines, allowing jobs to be swapped between lines. Additionally, we want to investigate how our approach generalizes to very large permutations, which result in a large action space, as well as to permutations of varying lengths. To achieve this, we plan to expand our dataset with additional real-world data and synthetically generated data. Furthermore, we plan to enhance our approach by incorporating additional operators beyond the swap operator. Finally, we want to combine our approach with various constructive heuristics that generate the initial permutations to potentially find even better overall solutions.